\documentclass[letterpaper, 10 pt, conference]{ieeeconf}  

\IEEEoverridecommandlockouts                              

\usepackage{cite}
\usepackage{amsmath,amssymb,amsfonts}
\usepackage{algorithmic}
\usepackage{booktabs}
\usepackage{multirow}
\usepackage{tabularx}
\usepackage{graphicx}
\usepackage{textcomp}
\usepackage{xcolor}
\newtheorem{theorem}{Theorem}

\newtheorem{definition}{Definition}

\usepackage{graphicx}
\title{\LARGE \bf
Universal Smoothness via Bernstein Polynomials: A Constructive Approximation Approach for Activation Functions
}

\author{Wentao Zhang$^{1}$, Yutong Zhang$^{2}$, Yifan Zhu$^{2}$ and Wentao Mo$^{1}$
\thanks{$^{1}$Wentao Zhang and Wentao Mo are with Tsinghua Shenzhen International Graduate School, Tsinghua University, Shenzhen, China {\tt\small \{zhang-wt24, 
mow10\}@mails.tsinghua.edu.cn}}%
\thanks{$^{2}$Yutong Zhang and Yifan Zhu  is with the College of Computer Science, Sichuan University, Chengdu, China
        {\tt\small zyfan@stu.scu.edu.cn}}%
}

\begin{document}

\maketitle
\thispagestyle{empty}
\pagestyle{empty}

\begin{abstract}

The efficacy of deep neural networks is heavily reliant on the design of non-linear activation functions, yet existing approaches often struggle to balance optimization stability with computational efficiency. While piecewise linear functions offer inference speed, they suffer from optimization instability due to non-differentiability at the origin, whereas smooth counterparts typically incur significant computational overhead through their reliance on transcendental operations. To address these limitations, this paper proposes a general smoothing framework based on constructive approximation theory and introduces the Bernstein Linear Unit (BerLU). This novel activation function utilizes Bernstein polynomials to construct a differentiable quadratic transition region that effectively eliminates singularities while maintaining a piecewise linear structure. Theoretical analysis demonstrates that the proposed method guarantees strictly continuous differentiability and a non-expansive Lipschitz constant of one, which ensures stable gradient propagation and prevents the gradient explosion problems common in deep architectures. Comprehensive empirical evaluations across representative Vision Transformer and Convolutional Neural Network architectures confirm that this approach consistently outperforms state-of-the-art baselines on standard image classification benchmarks while delivering superior computational and memory efficiency.

\end{abstract}

\section{INTRODUCTION}

The immense success of Deep Neural Networks (DNNs) in fields such as computer vision and natural language processing is largely attributed to the design of non-linear activation functions, which empower networks to model complex data distributions\cite{dubey2022activation}. In early research, saturating activation functions like Sigmoid and Tanh\cite{tanh} dominated, but they were prone to the vanishing gradient problem in deep networks. This landscape was fundamentally shifted by the proposal of the Rectified Linear Unit (ReLU)\cite{hahnloser2000digital,jarrett2009best,nair2010rectified}. With its sparsity resulting from one-sided suppression and non-saturating nature in the positive domain, ReLU significantly alleviated the vanishing gradient problem and accelerated model convergence, becoming the standard for modern deep learning architectures.

However, ReLU is not without flaws. Its limitations are primarily manifested in two aspects. The first is the Dying ReLU problem. Since the gradient of ReLU in the negative interval is constantly zero, once a neuron enters a negative activation state, its weights can no longer be updated, leading to a waste of network capacity. To mitigate this, variants like Leaky ReLU\cite{maas2013rectifier} and Parameterized Rectified Linear Unit (PReLU)\cite{prelu} were proposed, which introduce a small slope in the negative interval to maintain gradient flow. The second limitation is non-smoothness. ReLU and its variants are non-differentiable at the origin. Recent theoretical studies indicate that the smoothness of activation functions is critical for the training dynamics of deep networks\cite{hayou2019impact,biswas2022smooth,zhang2025general}. Smooth activation functions can induce flatter loss landscapes, thereby making it easier for gradient-based optimization algorithms to escape local minima and improving generalization\cite{li2018visualizing}. In pursuit of smoothness, researchers have proposed novel activation functions such as Gaussian Error Linear Unit (GELU)\cite{gelu}, Swish (SiLU)\cite{silu} , and Mish\cite{mish}. These functions achieve smooth transitions by introducing Gaussian error functions or exponential functions  and have demonstrated superior performance in large-scale models like BERT\cite{devlin2019bert} and GPT-3\cite{brown2020language}. However, these functions rely on transcendental functions, which incur non-negligible computational overhead in large-scale training and inference compared to ReLU, which employs simple thresholding operations.

Consequently, developing an ideal activation function requires simultaneously addressing two critical issues: the Dying ReLU problem and the optimization instability caused by non-smoothness. First, to guarantee global gradient flow and prevent neuron death, it is essential to retain the piecewise linear structure with a non-zero negative slope, characteristic of the Leaky ReLU family. Second, to induce a flatter loss landscape and improve generalization, the non-differentiable singularity at the origin must be eliminated. While existing smooth functions like GELU address the latter, they rely on computationally expensive transcendental functions. To achieve both objectives efficiently, this paper proposes a novel approach that mollifies the Leaky ReLU structure. Inspired by the constructive approximation capabilities of Bernstein Polynomials, we introduce a general activation function smoothing framework. By applying this framework to smooth the non-differentiable transition of Leaky ReLU, we derive a new activation function: the Bernstein Linear Unit (BerLU). The main contributions of this paper are summarized as follows:
\begin{itemize}
\item \textbf{General Applicability:} We present a generalized smoothing framework grounded in constructive approximation theory using Bernstein polynomials. This flexible tool systematically transforms non-smooth piecewise linear activations into strictly differentiable counterparts while preserving their fundamental geometric properties.

\item \textbf{High Computational Efficiency:} We propose the Bernstein Linear Unit, named BerLU, which replaces the singularity of Leaky ReLU with a quadratic Bernstein polynomial transition. Unlike approximations relying on expensive transcendental operations like GELU, BerLU employs basic arithmetic to significantly reduce computational overhead during training and inference while maintaining optimization stability.

\item \textbf{Theoretical Guarantees:} We provide a rigorous theoretical analysis leveraging the variation-diminishing property of Bernstein polynomials. BerLU guarantees smoothness and a non-expansive Lipschitz constant of 1. This foundation ensures continuous gradient flow and facilitates flatter loss landscapes, thereby stabilizing the optimization dynamics of deep neural networks.

\item \textbf{Extensive Empirical Validation:} We conduct comprehensive experiments across representative Vision Transformers (ViT, DeiT, TNT) and ConvNeXt on CIFAR and ImageNet benchmarks. The results demonstrate that BerLU consistently outperforms state-of-the-art baselines, notably surpassing GELU by 8.4\% on CIFAR-100, establishing itself as a robust and architecture-agnostic solution for large-scale vision tasks.
\end{itemize}
\section{RELATED WORK}
Existing activation functions generally fall into two primary categories: parametric piecewise linear functions designed for efficiency, and smooth transcendental functions designed for optimization stability.
\subsection{Parametric and Adaptive Linear Units}
The Rectified Linear Unit (ReLU) \cite{hahnloser2000digital,jarrett2009best,nair2010rectified} remains the cornerstone of deep learning due to its computational sparsity and alleviation of the vanishing gradient problem. However, its strict zero-output constraint for negative inputs leads to the dying ReLU phenomenon. To address this, Leaky ReLU \cite{maas2013rectifier} introduces a small, fixed negative slope, ensuring that gradients can propagate globally even through inactive neurons. PReLU \cite{prelu} extends this by formulating the negative slope as a learnable parameter, enabling the network to adaptively optimize the activation shape for specific tasks. While these variants successfully restore information flow, they share a fundamental structural limitation: non-differentiability at the origin. This singularity creates a kink in the loss landscape, which can hinder the convergence of optimization algorithms and limit the exploitation of higher-order curvature information during training.

\subsection{Gating and Transcendental Smooth Functions}
Beyond piecewise linear modifications, a distinct line of research has focused on smooth activation functions constructed via transcendental operations to address distributional bias and non-differentiability. The Exponential Linear Unit (ELU) \cite{elu} was introduced to yield smooth values in the negative domain, effectively counteracting mean-shift phenomena—a concept further streamlined by the Continuously Differentiable Exponential Linear Unit (CELU) \cite{celu} to simplify parameter tuning. In recent years, the preference for smooth activations has intensified due to their superior theoretical properties in optimization. For instance, Swish (SiLU) \cite{silu} utilizes a fully differentiable self-gated profile to stabilize training dynamics and enhance gradient flow. Similarly, Mish \cite{mish} facilitates effective signal propagation in deep networks and improves generalization by uniquely combining unbounded positive responses with controlled, smooth negative saturation. In the realm of applied deep learning, GELU \cite{gelu} and Swish-based variants have established themselves as the prevailing activation functions for large-scale vision and language architectures. They serve as the standard configuration in models such as BERT and RoBERTa \cite{devlin2019bert} and are utilized in ViT \cite{dosovitskiy2020vit}. Furthermore, they are widely considered to be the underlying mechanism supporting the GPT series \cite{radford2019gpt2,achiam2023gpt}, as well as contemporary systems like PaLM \cite{chowdhery2023palm}, LLaMA \cite{touvron2023llama}, and DeepSeek-V2 \cite{liu2024deepseek}. This widespread adoption highlights the critical necessity of smooth activations for achieving stability and scalability in deep learning. At the same time, the selection of SiLU by certain open-source models, most notably Mixtral \cite{jiang2024mixtral}, indicates that the investigation into alternative activation functions remains active.

\section{METHODOLOGY}
\subsection{Theoretical Motivation}
According to theoretical insights from Hayou \cite{hayou2019impact}, the smoothness of activation functions is a determinant factor in preserving signal fidelity across deep networks. The analysis reveals that for smooth activations, the inter-layer neuron correlation $c_l$ approaches unity at a relatively slow rate of $O(1/l)$, described by the relation $1 - c_l \sim \beta_q/l$. Here, $\beta_q$ depends on the target variance $q$ and the specific function, while $l$ indicates depth. Conversely, non-smooth functions such as ReLU exhibit a much faster correlation decay of $O(1/l^2)$, following the asymptotic behavior $1 - c_l \sim 9\pi^2/2l^2$. This significant theoretical divergence highlights how non-differentiability at the origin creates a bottleneck that disrupts gradient stability and hampers effective information flow.
\subsection{Bernstein-Mollified Adaptive Activation}
To address the optimization challenges posed by the non-differentiability of ReLU, we propose the Bernstein Linear Unit (BerLU). Unlike heuristic smoothing approaches, BerLU is constructed based on constructive approximation theory, leveraging the variation-diminishing property of Bernstein polynomials to create a strictly continuous transition between the inactive and active regimes of the rectifier. Formally, we seek a smoothing function $f(x)$ defined on a transition interval $\mathcal{I}_\epsilon = [-\epsilon, \epsilon]$, where $\epsilon > 0$ is a hyperparameter controlling the mollification radius. To bridge the negative linear segment parameterized with slope $\alpha$ and the positive identity mapping, we employ a probabilistic linear combination of Bernstein basis polynomials. Let $t: [-\epsilon, \epsilon] \to [0, 1]$ be the affine coordinate mapping defined as $t(x) = \frac{x + \epsilon}{2\epsilon}$. A Bernstein polynomial of degree $n$ is given by:
$$B_n(t) = \sum_{k=0}^{n} \beta_k b_{k,n}(t), \quad \text{where } b_{k,n}(t) = \binom{n}{k} t^k (1-t)^{n-k}$$
Here, $\{\beta_k\}_{k=0}^n$ represents the set of control points that dictate the geometric shape of the transition curve. To balance computational efficiency with smoothness, we adopt a degree-$2$ approximation ($n=2$), which corresponds to a quadratic Bézier curve. The activation function $f(x)$ is thus defined piecewise:
$$f(x) =
\begin{cases}
\alpha x, & x < -\epsilon \\
\sum_{k=0}^{2} \beta_k b_{k,2}\left(\frac{x+\epsilon}{2\epsilon}\right), & -\epsilon \le x \le \epsilon \\
x, & x > \epsilon
\end{cases}$$
The control coefficients $\beta_0, \beta_1, \beta_2$ are not arbitrary; they are uniquely determined by imposing $C^0$ (continuity) and $C^1$ (differentiability) constraints at the boundaries $x = -\epsilon$ and $x = \epsilon$. First, matching the value and derivative of the Leaky component ($\alpha x$) at the left boundary ($t=0$) determines the first two control points. Specifically, the value condition gives $\beta_0 = -\alpha \epsilon$. The derivative condition, accounting for the domain scaling factor $\frac{1}{2\epsilon}$, requires $\frac{1}{\epsilon}(\beta_1 - \beta_0) = \alpha$, which simplifies to $\beta_1 = 0$. Then, at the right boundary ($t=1$), matching the identity function ($x$) sets $\beta_2 = \epsilon$. Substituting these values into the derivative equation at the right boundary confirms a slope of 1, ensuring a smooth transition to the linear region. By substituting the solved coefficients $\boldsymbol{\beta} = [-\alpha\epsilon, 0, \epsilon]$ back into the Bernstein expansion and simplifying the terms, we obtain the computationally efficient form of BerLU:
$$\text{BerLU}(x) =
\begin{cases}
\alpha x, & x < -\epsilon \\
\frac{1-\alpha}{4\epsilon}x^2 + \frac{1+\alpha}{2}x + \frac{(1-\alpha)\epsilon}{4}, & -\epsilon \le x \le \epsilon \\
x, & x > \epsilon
\end{cases}$$
This gives us the analytical form of the BerLU activation function. The gradient with respect to the input $x$ is explicitly given by:
$$\frac{\partial f}{\partial x} =
\begin{cases}
\alpha, & x < -\epsilon \\
\frac{1-\alpha}{2\epsilon}x + \frac{1+\alpha}{2}, & -\epsilon \le x \le \epsilon \\
1, & x > \epsilon
\end{cases}$$
The transition region ($-\epsilon \le x \le \epsilon$) provides a linear gradient interpolation. This introduces non-zero second-order derivatives, creating a smoother optimization landscape compared to piecewise linear activations. Furthermore, by optimizing $\alpha$, the network essentially adjusts the control point $\beta_0$, dynamically reshaping the activation geometry to fit the data distribution.
\subsection{Lipschitz Continuity Analysis}
Lipschitz continuity has emerged as a cornerstone in deep learning research, crucial for constructing networks that demonstrate enhanced reliability, robustness, and generalization capabilities. Bridging theoretical frameworks and practical deployment, this property enforces a stricter constraint than mere continuity by placing an upper bound on the function's rate of variation \cite{lip1,sun2025entropy,lip4}.

\begin{definition}
A function $f : \mathbb{R} \to \mathbb{R}$ is said to be Lipschitz continuous if there exists a constant $C \ge 0$ such that for all $x, y \in \text{dom}(f)$, the inequality $|f(x) - f(y)| \le C|x - y|$ holds. The smallest value $C$ that satisfies this condition is defined as the Lipschitz constant, denoted as $L$.
\end{definition}

\begin{theorem}
The proposed BerLU activation function is Lipschitz continuous with the constant $L_{\text{BerLU}} = \max(1, |\alpha|)$.
\end{theorem}

The Lipschitz constant $L$ is defined as the global supremum of the derivative magnitude $|f'(x)|$. Outside the transition region ($|x| > \epsilon$), the derivative takes constant values of either $\alpha$ or $1$. Within the transition region ($|x| \le \epsilon$), since BerLU is constructed as a quadratic polynomial, its derivative $f'(x)$ is a linear function. Due to the monotonicity of linear functions, the maximum absolute value of the derivative on the closed interval $[-\epsilon, \epsilon]$ must occur at one of the boundaries. Consequently, the global maximum is strictly bounded by the larger of the two endpoint slopes, yielding $L_{\text{BerLU}} = \max(1, |\alpha|)$. In practice, the negative slope parameter typically satisfies $|\alpha| \ll 1$ (e.g., $\alpha=0.01$), implying that the Lipschitz constant of BerLU is exactly $1$.

To demonstrate a critical structural superiority, we evaluate BerLU against prevailing smooth activation functions through the rigorous lens of Lipschitz continuity. Detailed analysis reveals that standard smooth activations typically function as expansive operators, characterized by Lipschitz constants exceeding unity. Specifically, widely adopted functions such as GELU ($L \approx 1.084$), SiLU ($L \approx 1.100$), and Mish ($L \approx 1.089$) exhibit maximum derivative magnitudes greater than one. In the context of deep architectures, where backpropagation is mathematically governed by the chain rule across a sequence of layer Jacobians, such expansivity ($L > 1$) is non-trivial. It implies that the upper bound of the gradient norm can undergo exponential growth relative to network depth $d$, theoretically scaling as $O(L^d)$. This multiplicative amplification creates a inherent structural vulnerability, potentially driving gradient explosion or inducing high volatility during the optimization of extremely deep models. Conversely, BerLU is engineered to strictly enforce $L=1.000$, classifying it as a strictly non-expansive operator. By constraining the derivative within the unit interval, BerLU ensures that the activation layer never intrinsically amplifies the error signal or gradient norm ($|\nabla f(x)| \le |\nabla x|$). Consequently, BerLU provides a quantifiable theoretical safeguard, serving as a vital stabilizer for gradient flow while simultaneously preserving the differentiability benefits required for efficient convergence.
\begin{table*}[t]
    \centering
    \caption{Test Accuracy of ViT, DeiT, and TNT Architectures on CIFAR-10, CIFAR-100, and ImageNet-1K Benchmarks over a 100-Epoch Training Course.}
    \label{tab:test_accuracy}
    \begin{tabular*}{\textwidth}{@{\extracolsep{\fill}} ll ccccccc}
    \toprule
    \multirow{2}{*}{\textbf{Dataset}} & \multirow{2}{*}{\textbf{Model}} & \multicolumn{7}{c}{\textbf{Activation Functions}} \\
    \cmidrule{3-9}
     & & \textbf{GELU} & \textbf{ELU} & \textbf{PReLU} & \textbf{CELU} & \textbf{SiLU} & \textbf{Mish} & \textbf{BerLU} \\
    \midrule
    \multirow{4}{*}{CIFAR-10} 
      & ViT-Tiny & 70.4 $\pm$ 0.2 & 66.4 $\pm$ 0.5 & 78.0 $\pm$ 0.6 & 66.5 $\pm$ 0.6 & 68.6 $\pm$ 0.3 & 68.7 $\pm$ 0.3 & \textbf{78.5 $\pm$ 0.2} \\
      & DeiT-Tiny & 72.4 $\pm$ 0.7 & 67.6 $\pm$ 0.6 & 75.4 $\pm$ 0.1 & 67.7 $\pm$ 0.8 & 69.9 $\pm$ 0.5 & 70.2 $\pm$ 0.6 & \textbf{77.6 $\pm$ 0.3} \\
      & TNT-Small & 73.7 $\pm$ 0.5 & 69.5 $\pm$ 0.6 & 75.8 $\pm$ 0.3 & 68.7 $\pm$ 0.2 & 71.1 $\pm$ 0.7 & 71.6 $\pm$ 0.8 & \textbf{77.2 $\pm$ 0.5} \\
    \cmidrule{2-9}
      & Average & 72.2 $\pm$ 0.5 & 67.8 $\pm$ 0.6 & 76.4 $\pm$ 0.3 & 67.6 $\pm$ 0.5 & 69.9 $\pm$ 0.5 & 70.2 $\pm$ 0.6 & \textbf{77.8 $\pm$ 0.3} \\
    \midrule
    \multirow{4}{*}{CIFAR-100} 
      & ViT-Tiny & 32.6 $\pm$ 0.8 & 28.9 $\pm$ 0.1 & 43.2 $\pm$ 1.0 & 28.9 $\pm$ 0.2 & 31.2 $\pm$ 0.6 & 30.6 $\pm$ 0.8 & \textbf{45.5 $\pm$ 0.3} \\
      & DeiT-Tiny & 46.6 $\pm$ 0.9 & 56.9 $\pm$ 0.0 & 50.0 $\pm$ 0.5 & 40.5 $\pm$ 0.5 & 43.5 $\pm$ 0.6 & 43.8 $\pm$ 1.0 & \textbf{53.5 $\pm$ 0.6} \\
      & TNT-Small & 47.5 $\pm$ 0.8 & 43.6 $\pm$ 0.3 & 49.0 $\pm$ 0.7 & 43.0 $\pm$ 0.5 & 45.0 $\pm$ 0.9 & 45.5 $\pm$ 0.8 & \textbf{52.7 $\pm$ 0.5} \\
    \cmidrule{2-9}
      & Average & 42.2 $\pm$ 0.8 & 43.1 $\pm$ 0.1 & 47.4 $\pm$ 0.7 & 37.5 $\pm$ 0.4 & 39.9 $\pm$ 0.7 & 40.0 $\pm$ 0.9 & \textbf{50.6 $\pm$ 0.5} \\
    \midrule
    \multirow{3}{*}{ImageNet-1K} 
      & ViT-Tiny & 53.9 $\pm$ 0.3 & 37.2 $\pm$ 0.6 & 56.8 $\pm$ 0.3 & 37.6 $\pm$ 0.5 & 46.1 $\pm$ 0.7 & 46.9 $\pm$ 1.1 & \textbf{57.3 $\pm$ 0.2} \\
      & DeiT-Tiny & 61.7 $\pm$ 0.4 & 49.1 $\pm$ 0.7 & 60.8 $\pm$ 0.4 & 48.9 $\pm$ 0.8 & 58.5 $\pm$ 0.7 & 58.9 $\pm$ 0.3 & \textbf{63.8 $\pm$ 0.6} \\
    \cmidrule{2-9}
      & Average & 57.8 $\pm$ 0.4 & 43.2 $\pm$ 0.7 & 58.8 $\pm$ 0.4 & 43.3 $\pm$ 0.7 & 52.3 $\pm$ 0.7 & 52.9 $\pm$ 0.7 & \textbf{60.6 $\pm$ 0.4} \\
    \bottomrule
    \end{tabular*}
\end{table*}

\begin{table*}[t] 
\caption{Test Accuracy of ConvNeXt on CIFAR-10, CIFAR-100, and ImageNet-1K Benchmarks over a 100-Epoch Training Course.}
\label{table 4}
\centering
\begin{tabular*}{\textwidth}{@{\extracolsep{\fill}} ll ccccccc}
\toprule
\multirow{2}{*}{\textbf{Dataset}} & \multirow{2}{*}{\textbf{Model}} & \multicolumn{7}{c}{\textbf{Activation Functions}} \\
\cmidrule{3-9} 
 & & \textbf{GELU} & \textbf{ELU} & \textbf{PReLU} & \textbf{CELU} & \textbf{SiLU} & \textbf{Mish} & \textbf{BerLU} \\
\midrule
CIFAR-10    & ConvNeXt & 64.9 $\pm$ 0.4 & 59.8 $\pm$ 0.5 & 64.6 $\pm$ 1.4 & 59.8 $\pm$ 0.5 & 60.6 $\pm$ 0.2 & 61.4 $\pm$ 0.4 & \textbf{66.5 $\pm$ 0.2} \\
CIFAR-100   & ConvNeXt & 36.6 $\pm$ 0.3 & 30.3 $\pm$ 0.4 & 35.2 $\pm$ 0.5 & 30.5 $\pm$ 0.2 & 35.0 $\pm$ 0.9 & 35.3 $\pm$ 0.7 & \textbf{37.9 $\pm$ 0.3} \\
ImageNet-1K & ConvNeXt & 72.9 $\pm$ 0.3 & 71.7 $\pm$ 0.5 & 72.9 $\pm$ 0.5 & 71.8 $\pm$ 0.9 & 72.3 $\pm$ 0.7 & 72.8 $\pm$ 0.6 & \textbf{73.5 $\pm$ 0.2} \\
\bottomrule
\end{tabular*}
\end{table*}
\section{EXPERRIMENTS}
\subsection{Experimental Settings}
\textbf{Datasets.} To comprehensively evaluate the effectiveness of the proposed BerLU in computer vision tasks, we conducted experiments on three standard benchmark datasets with increasing complexity: CIFAR-10, CIFAR-100 \cite{krizhevsky2009cifar}, and ImageNet-1K \cite{deng2009imagenet}. CIFAR-10 and CIFAR-100 serve as foundational benchmarks for small-scale image classification, consisting of $32 \times 32$ pixel images covering 10 and 100 categories, respectively. To assess performance on large-scale classification, we utilize the ImageNet-1K dataset, which contains high-resolution images across 1,000 diverse categories, providing a robust testbed for generalization capability.

\textbf{Baselines.} We benchmark BerLU against a comprehensive set of representative activation functions, including piecewise linear variants and smooth transcendental functions. The baseline methods include  GELU\cite{gelu}, ELU\cite{elu}, PReLU\cite{prelu}, CELU\cite{celu}, SiLU\cite{silu}, and Mish\cite{mish}. This diverse comparison group allows us to verify whether BerLU combines the efficiency of linear units with the optimization stability of smooth units

\textbf{Implementation Details.} The experimental evaluation was performed on a high-performance cluster leveraging eight NVIDIA A100 GPUs. Optimization was driven by the AdamW algorithm with a configured weight decay of 0.05 for regularization. To maintain training stability, we applied a gradient clipping threshold of 1.0, while model weights were initialized via a truncated normal distribution. The learning process minimized the cross-entropy objective using a cosine annealing schedule integrated with a linear warmup period.

\textbf{Hyperparameters.} For the proposed BerLU activation, we set the smoothing parameter $\epsilon$, which corresponds to the mollification radius, to a default value of $10^{-2}$. The rationale for this selection is discussed in detail in Section \ref{D}, Sensitivity Analysis. Unlike $\epsilon$, the coefficient $\alpha$ is designed as a learnable parameter rather than a fixed scalar; it is initialized to $0.01$ and optimized jointly with the model weights during training. Each experimental configuration was executed three times, and we report the mean and standard deviation to ensure statistical reliability.
\begin{table*}[!t] 
\centering
\caption{Computational Efficiency Analysis of ViT on CIFAR-10: Forward/Backward Latency and Peak Memory Footprint.}
\label{jisuan}
\begin{tabular*}{\textwidth}{@{\extracolsep{\fill}}llccccccc}
\toprule
\multirow{2}{*}{\textbf{Computational Cost}} & \multirow{2}{*}{\textbf{Model}} & \multicolumn{7}{c}{\textbf{Activation Functions}} \\
\cmidrule{3-9}
& & \textbf{GELU} & \textbf{ELU} & \textbf{PReLU} & \textbf{CELU} & \textbf{SiLU} & \textbf{Mish} & \textbf{BerLU} \\
\midrule
Forward Time (ms)  & ViT & 59.2  & 57.7  & 57.5 & 53.3  & 54.2  & 56.2  & 55.8 \\
Backward Time (ms) & ViT & 148.0 & 155.8 & 156.5 & 154.9 & 154.6 & 148.8 &  154.2\\
Peak Memory (GB)   & ViT & 11.53 & 11.07 & 9.50 & 11.06 & 12.25 & 11.05 & 10.58 \\
\bottomrule
\end{tabular*}
\end{table*}
\subsection{Evaluation on Vision Transformers}
To rigorously assess the expressive power of our proposed activation function, we executed a comprehensive suite of experiments across three distinct Vision Transformer variants: ViT \cite{dosovitskiy2020vit}, DeiT \cite{touvron2021training}, and TNT \cite{han2021transformer}. This diverse assortment was chosen to encompass standard self-attention paradigms, knowledge distillation strategies, and fine-grained feature interaction mechanisms, respectively. In terms of data evaluation, we leveraged CIFAR-10 and CIFAR-100 to probe the robustness of BerLU across varying distribution shifts and class granularities. Additionally, the ImageNet-1K benchmark was utilized to validate performance in high-complexity scenarios, benefiting from its high-resolution imagery and extensive categorical diversity.

The quantitative results in Table I demonstrate that BerLU consistently outperforms both non-parametric functions (e.g., GELU) and learnable baselines (e.g., PReLU) across all tested architectures. On CIFAR-10, BerLU achieves top-tier accuracy, improving upon the strong PReLU baseline. This advantage is most prominent on the challenging CIFAR-100 dataset, where BerLU achieves an average accuracy of 50.6\%, outperforming PReLU by 3.2\% and the standard GELU by 8.4\%. This superiority extends to ImageNet-1K, where BerLU maintains the highest average top-1 accuracy of 60.6\%, verifying its scalability to large-scale data regimes.These results validate two critical design choices. First, the consistent improvement over PReLU indicates that the  smoothness provided by Bernstein polynomials ensures better gradient flow than piecewise linear functions. Second, the performance gap over GELU highlights the benefit of a learnable shape parameter, which allows the model to adaptively optimize the activation curvature for different layers rather than relying on a fixed, deterministic form.

\subsection{Evaluation on Convolutional Neural Networks}
To demonstrate the architecture-agnostic nature of BerLU, we extended our evaluation to the convolutional domain using ConvNeXt \cite{liu2022convnet}, a state-of-the-art CNN constructed with modern design principles. Integrating BerLU into this high-performance framework serves as a rigorous stress test. It allows us to determine whether the observed performance gains are strictly intrinsic to attention-based mechanisms or if they genuinely transcend specific inductive biases to generalize across diverse neural architectures.

As shown in Table II, BerLU consistently outperforms baselines, achieving top-tier accuracies of 66.5\%, 37.9\%, and 73.5\% on CIFAR-10, CIFAR-100, and ImageNet-1K, respectively. The performance advantage is particularly evident against the widely adopted SiLU and GELU. Specifically, on the fine-grained CIFAR-100 benchmark, BerLU surpasses SiLU (35.0\%) and GELU (36.6\%) by significant margins of 2.9\% and 1.3\%, respectively. Even on the large-scale ImageNet-1K dataset, BerLU maintains a clear lead over PReLU (72.9\%) and GELU (72.9\%), demonstrating superior scalability. This superiority over PReLU confirms that eliminating the singularity at the origin is essential for gradient stability, while the advantage over GELU and SiLU suggests that BerLU's polynomial transition preserves feature magnitude better than aggressive self-gating mechanisms. These results establish BerLU as an architecture-agnostic activation effective for both CNNs and Transformers.
\begin{figure}[t]
    \centering
    \includegraphics[width=\linewidth]{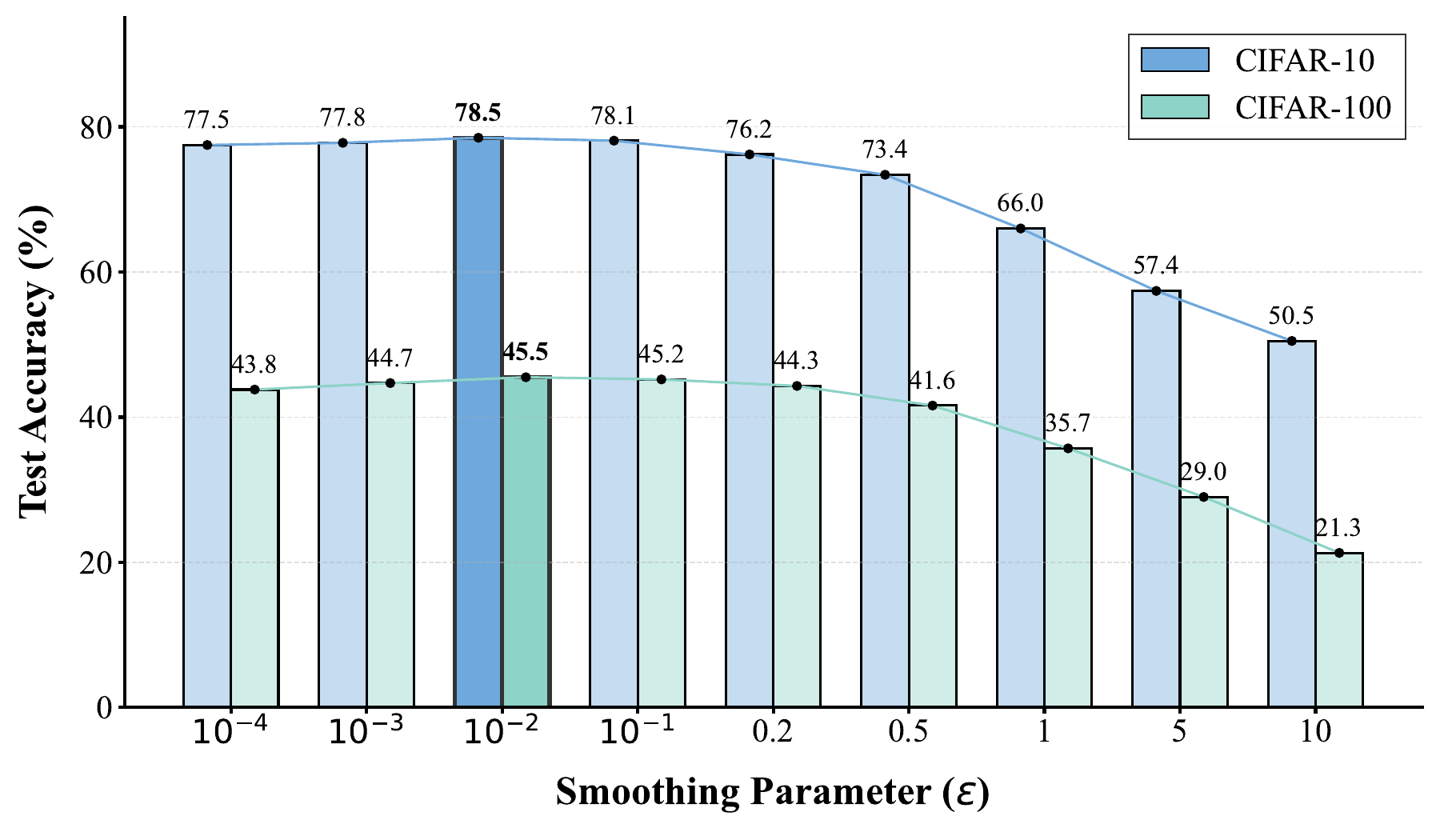}
    \caption{Impact of the Smoothing Parameter $\epsilon$ on ViT Classification Performance across CIFAR Datasets trained for 100 Epochs.}
    \label{sen}
\end{figure}

\subsection{Sensitivity Analysis}
\label{D}
This subsection analyzes the sensitivity of classification accuracy to the smoothing parameter $\epsilon$. We employed the ViT-Tiny architecture on CIFAR-10 and CIFAR-100 benchmarks, sweeping $\epsilon$ across the discrete set $\{10^{-4}, 10^{-3}, 10^{-2}, 10^{-1}, 0.2, 0.5, 1, 5, 10\}$. To ensure statistical reliability, each hyperparameter setting was evaluated through three independent trials. The resulting mean accuracy and standard deviation are visualized in Fig. \ref{sen}.

    
    

The accuracy exhibits a rise-then-fall trend, peaking at $\epsilon = 10^{-2}$ with top-1 accuracies of 78.5\% on CIFAR-10 and 45.5\% on CIFAR-100. Performance remains highly stable for small $\epsilon \in [10^{-4}, 10^{-1}]$, where the accuracy fluctuation is negligible (within 1.5\%), demonstrating the method's robustness to hyperparameter selection. However, increasing $\epsilon$ beyond 0.2 causes significant degradation. Specifically, at $\epsilon=10$, the accuracy plummets to 50.5\% on CIFAR-10 and, more drastically, to 21.3\% on CIFAR-100. This severe drop on the more complex CIFAR-100 dataset indicates that while moderate smoothing benefits optimization, excessive smoothing effectively linearizes the activation function, eroding the non-linearity essential for discriminating fine-grained features. Thus, $\epsilon = 10^{-2}$ strikes the optimal balance between gradient stability and representational capacity.
\subsection{Computational Efficiency Analysis}
To assess the practical deployment feasibility of BerLU compared to established norms, we performed a comprehensive computational efficiency analysis. Key performance metrics, specifically the average latency for forward and backward propagation and maximum memory consumption, were recorded. The quantitative results are presented in Table \ref{jisuan}.

Regarding forward propagation, BerLU achieves a latency of 55.8 ms, surpassing the standard GELU at 59.2 ms. This efficiency stems from its reliance on simple polynomial operations rather than complex transcendental functions. Notably, BerLU demonstrates superior memory efficiency by consuming only 10.58 GB of peak memory, significantly lower than GELU (11.53 GB) and SiLU (12.25 GB). While its backward pass time of 154.2 ms is slightly higher than GELU (148.0 ms), it remains comparable to PReLU and CELU. This marginal overhead is acceptable given the substantial improvements in inference speed and memory usage, confirming that BerLU effectively balances computational efficiency with optimization smoothness.

\section{CONCLUSIONS}

In this paper, we presented the Bernstein Linear Unit (BerLU), a novel activation function grounded in constructive approximation theory, designed to resolve the persistent trade-off between computational efficiency and optimization stability in deep neural networks. By leveraging Bernstein polynomials to mollify the Leaky ReLU structure, BerLU effectively eliminates the non-differentiable singularity at the origin without relying on computationally expensive transcendental operations.Theoretical analysis confirms that BerLU guarantees smoothness and a non-expansive Lipschitz constant of one, providing a rigorous guarantee for stable gradient propagation in deep architectures. Extensive empirical evaluations across representative Vision Transformers and ConvNeXt architectures demonstrate that BerLU consistently outperforms state-of-the-art baselines on CIFAR and ImageNet benchmarks. Notably, compared to the widely adopted GELU, BerLU delivers superior generalization performance while maintaining lower memory footprint and faster inference latency. Consequently, BerLU stands as an architecture-agnostic, efficient, and robust solution, offering significant potential to serve as a standard component for large-scale computer vision tasks.





\bibliographystyle{IEEEtran}

\bibliography{1}


\end{document}